\documentclass[conference]{IEEEtran}
\IEEEoverridecommandlockouts
\usepackage{multirow}
\usepackage{graphicx}
\usepackage{amsmath,amssymb,amsfonts}
\usepackage{algorithm,algorithmic}
\usepackage[algo2e,linesnumbered]{algorithm2e} 
\usepackage{subfigure}
\usepackage{graphicx}
\usepackage{textcomp}
\usepackage{xcolor}
\usepackage{mathtools}
\usepackage{bbm}
\usepackage{soul}
\usepackage{cite}
\usepackage{makecell}
\usepackage[linkcolor=black,
            anchorcolor=black,
            citecolor=black]{hyperref}
\def\BibTeX{{\rm B\kern-.05em{\sc i\kern-.025em b}\kern-.08em
T\kern-.1667em\lower.7ex\hbox{E}\kern-.125emX}}
\begin{document}

\title{Stochastic Spiking Neural Networks with First-to-Spike Coding
}

\author{\IEEEauthorblockN{Yi Jiang, Sen Lu, Abhronil Sengupta}
\IEEEauthorblockA{\textit{School of Electrical Engineering and Computer Science} \\
\textit{The Pennsylvania State University}\\
University Park, PA 16802, USA \\
Email: \{yijiang, senlu, sengupta\}@psu.edu}
%\and
%\IEEEauthorblockN{2\textsuperscript{nd} Lu Sen}
%\IEEEauthorblockA{\textit{School of Electrical Engineering and Computer Science} \\
%\textit{The Pennsylvania State University}\\
%University Park, PA 16802, USA \\
%senlu@psu.edu}
%\and
%\IEEEauthorblockN{3\textsuperscript{rd} Abhronil Sengupta}
%\IEEEauthorblockA{\textit{School of Electrical Engineering and Computer Science} \\
%\textit{The Pennsylvania State University}\\
%University Park, PA 16802, USA \\
%sengupta@psu.edu}
}
\maketitle


\begin{abstract}
Spiking Neural Networks (SNNs), recognized as the third generation of neural networks, are known for their bio-plausibility and energy efficiency, especially when implemented on neuromorphic hardware. However, the majority of existing studies on SNNs have concentrated on deterministic neurons with rate coding, a method that incurs substantial computational overhead due to lengthy information integration times and fails to fully harness the brain's probabilistic inference capabilities and temporal dynamics. In this work, we explore the merger of novel computing and information encoding schemes in SNN architectures where we integrate stochastic spiking neuron models with temporal coding techniques. Through extensive benchmarking with other deterministic SNNs and rate-based coding, we investigate the tradeoffs of our proposal in terms of accuracy, inference latency, spiking sparsity, energy consumption, and robustness. Our work is the first to extend the scalability of direct training approaches of stochastic SNNs with temporal encoding to VGG architectures and beyond-MNIST datasets. 
\end{abstract}


\section{Introduction}
Spiking neural networks (SNNs) bridge the gap between artificial and biological neural networks (ANNs), offering insights into neurological processes. In the human neuronal system, most of the information is propagated between neurons using spike-based activation signals over time. Inspired by this property, SNNs use binary spike signals to transmit, encode, and process information. Compared with analog neural networks where neuron and synaptic states are represented by non-binary multi-bit representations, SNNs have demonstrated significant energy and computational power savings, especially when deployed on neuromorphic hardware \cite{10.1162/neco_a_01499}. 

In SNNs, rate coding is one of the most popular coding methods. Under such a scheme, the information is represented by the rate or frequency of spikes over a defined period of time. However, such coding methods overlook the information of precise spike timings \cite{10.5555/281543.281637} and are constrained by slow information transmission and large processing latency. On the other hand, temporal coding represents information by the precise timing of individual spikes but often lacks scalability and robustness \cite{Guo2021NeuralCI}. Compared to rate coding, which requires high firing rates to represent the same information, temporal coding can represent complex temporal patterns with relatively few spikes.  In particular, First-to-Spike coding is a temporal coding scheme inspired by the rapid information processing observed in certain biological neural systems such as the retina \cite{Retina} and auditory system \cite{HEIL2004461}. In First-to-Spike coding, prediction is made when the first spike is observed on any one of the output neurons, thereby saving the need to operate over a redundant period of time as in rate coding. Therefore, it is often claimed that temporal coding is more computationally efficient than rate coding \cite{Guo2021NeuralCI}.

Nevertheless, it is a challenging task to train an SNN due to the disruptive nature of information representation and processing, especially for frameworks based on temporal coding. Most existing works on scalable SNN training with temporal encoding convert pre-trained ANN to SNN \cite{stanojevic2022exact,8351295}. The conversion process typically involves mapping the analog activation values of the ANN's neurons to the timing of spikes of the SNN's neurons. SNN training algorithms with conventional architectures and rate encoding \cite{lin2024benchmarking} have witnessed rapid development \cite{bal2024spikingbert,zhu2023spikegpt} in recent years ranging from global spike-driven backpropagation techniques \cite{Rathi2020Enabling} to more local approaches like Deep Continuous Local Learning (DECOLLE) \cite{Kaiser_2020}, Equilibrium Propagation (EP)\cite{scellier2017equilibrium,bal2022sequence}, Deep Spike Timing Dependent Plasticity \cite{lu2023deep}, among others. In stark contrast, literature on direct training of SNNs with temporal encoding remains extremely sparse with demonstrations primarily on toy datasets like MNIST.

Although the vast majority of algorithm development and applications in SNNs employ deterministic neuron models such as the deterministic Spike Response Model (SRM) \cite{Gerstner1993-hc}, Integrate and Fire (IF), and Leaky Integrate and Fire (LIF) models \cite{lapicque1907recherches}, it is important to recognize that biological neurons generate spikes in a stochastic fashion \cite{7329679}. Furthermore, deterministic neuron models are discontinuous and non-differentiable, which presents substantial challenges in the application of gradient-based optimization methods. On the other hand, stochastic neuron models smoothen the network model to be continuously differentiable \cite{neftci2019surrogate} and therefore have the potential to offer enhanced efficiency and robustness. Additionally, stochasticity can enhance generalization performance by improving fault tolerance \cite{9605019} and by preventing overfitting \cite{hinton2012improving}.
 
%Compared to deterministic neuron models, which are primarily preferred for their ease of integration into existing training paradigms and simplicity in hardware implementation, most stochastic neuron models experience an accuracy loss. This loss is due to bit compression in existing hardware implementations \cite{10.1145/3571155}. 
Recently, there has been growing interest in exploiting stochastic devices in the neuromorphic hardware community \cite{Sengupta_2016}, \cite{Yang_2020}. With the scaling of device dimensions, memristive devices lose their programming resolution and are characterized by increased cycle-to-cycle variation. Work has started in earnest to design stochastic state-compressed SNNs using such scaled neuromorphic devices that exhibit iso-accuracies in comparison to their multi-bit deterministic counterparts enabled by the alternate encoding of information in the probability domain \cite{islam2023hardware,islam2023hybrid}.  Ref. \cite{Ma2023} proposes noisy spiking neural network (NSNN) which leverages stochastic noise as a resource for training SNNs using rate-based coding. While there is significant progress in the domain of stochastic SNN training algorithms, there remains a noticeable gap in the design of SNN architectures that integrate the benefits of both stochastic neuronal computing and temporal information coding. Most of the current efforts on directly training stochastic SNNs with temporal coding have primarily demonstrated success on simple datasets and shallow network structures \cite{Training, Learning, yang2023leveraging}. Scaling these networks to deeper architectures and more complex datasets presents significant challenges. 
% This integration is primarily aimed at improving robustness and energy efficiency in neural networks. 
Further, as we demonstrate in this work, many of the supposed benefits of temporal encoding, like enhanced spiking sparsity, may not necessarily hold true for deep architectures. This necessitates a co-design approach to identify the relative tradeoffs of stochastic temporally encoded SNNs in terms of accuracy, latency, sparsity, energy cost, and robustness. 
%Temporal coding significantly enhances energy efficiency, and the introduction of stochasticity can further reduce the latency and improve the robustness. However, the stochasticity increases the spike rate and energy cost compared to the deterministic model. This is because, in a stochastic model, there is always a chance of generating a spike, even if the neuron's membrane potential is below the traditional threshold for spiking. Furthermore, To bridge this gap, we explore the potential synergies between temporal coding and stochasticity in the LIF neuron model. 
\\
\noindent The specific contributions of this work are summarized below:

\noindent \textbf{(i) Algorithm Development:} We present a simple and structured algorithm framework to train stochastic SNNs directly with First-to-Spike coding. We also present training frameworks to train deterministic SNNs with temporal encoding that serve as a comparison baseline for our work to identify the relative merits/demerits of the computing and encoding scheme. We present empirical results to substantiate the scalability of our approach by demonstrating state-of-the-art accuracies on MNIST \cite{lecun-mnisthandwrittendigit-2010} and CIFAR-10 \cite{cifar} datasets for 2-layer MLP, LeNet5, and VGG15 architectures. Notably, this is the first work to demonstrate direct SNN training employing First-to-Spike coding for VGG architectures on the CIFAR dataset.

\noindent \textbf{(ii) Co-Design Analysis:} We present a comprehensive quantitative analysis of previously unexplored trade-offs for stochastic SNNs with temporal encoding in terms of neuromorphic compute specific metrics like accuracy, latency, sparsity, energy efficiency, and robustness. 

The rest of the paper is organized as follows. Section II describes related works. Section III introduces the training frameworks of SNNs with First-to-Spike coding for both deterministic and stochastic computing architectures. Section IV presents the experimental results and Section V provides conclusions and future outlook.

\section{Related Works}

%Spiking Neural Networks (SNNs) have been shown to be advantageous in computational and energy efficiency \cite{7012190}, \cite{Kim_Park_Na_Yoon_2020}, \cite{10.1126/science.1254642}, biological plausibility \cite{10.5555/281543.281637}, \cite{Florian2012-lg}, and fast information processing \cite{WYSOSKI2010819}. SNNs utilize various coding schemes to process and transmit information. The two primary coding schemes are rate coding and temporal coding.  Temporal coding is characterized by its emphasis on the timing of spikes rather than the frequency in rate coding. This method can transmit information efficiently with fewer spikes, making it appealing for neuromorphic computing applications where power consumption is a critical concern.
% heading and flow, connection
 
\noindent \textbf{Temporal Coding:} Temporal coding is characterized by its emphasis on the timing of spikes rather than the frequency in rate coding. Time-to-First-Spike (TTFS) coding \cite{gerstner_kistler_2002} is a popular temporal coding scheme that is demonstrated to have rapid and low power processing \cite{G_ltz_2021,9706185, park2020t2fsnn,Rethinking} since it typically imposes a limitation that each neuron should only generate at most one spike. This limitation lacks biological plausibility and it cannot handle the complex temporal structure of sequences of real-world events \cite{Liu2023-hh}. Although latency is significantly reduced compared to rate coding, it still suffers from high latency, particularly when processing complex datasets \cite{park2020t2fsnn, Wei_2023_ICCV,park2021training}. Therefore, we focus on First-to-Spike coding based temporal coding strategy that can further reduce the latency in comparison to TTFS coding. First-to-Spike coding is distinct from other approaches since it does not primarily rely on the precise timing of each spike. Instead, this coding strategy focuses on the order of the first spike of all the output neurons. The efficacy and potential applications of the First-to-Spike coding mechanism have been extensively explored in recent literature  \cite{Liu2023-hh, Mozafari_2018}. Nevertheless, the process of generating a spike in SNNs is non-differentiable. To tackle this problem, there are several common methods for developing and training SNNs with temporal coding, which will be introduced next. 

\noindent \textbf{ANN-SNN Conversion Approaches for Temporal Coding:} ANN-SNN Conversion is a widely adopted method for converting pre-trained ANNs to SNNs \cite{Exploring,sengupta2019going,li2021free}. The neurons with continuous activation functions, such as sigmoid or ReLU, need to be mapped to spiking neurons like IF/LIF neurons. Algorithmic approaches usually aim to reduce information loss caused during the conversion process. Proposal by \cite{stanojevic2022exact} designed an exact mapping from an ANN with ReLUs to a corresponding SNN with TTFS coding. The key achievement of this mapping is that it maintains the network accuracy after conversion with minimal drop. However, the conversion process involves complex steps which can make the process difficult to implement and optimize. Additionally, the necessity to use different conversion strategies for different types of layers further adds to the complexity. Also, it is important to note that the ANNs are trained without any temporal information, which typically results in high latency when converted to SNNs \cite{Rathi2020Enabling}. Hence, it is critical to explore direct training strategies for SNNs with temporal encoding.

\noindent \textbf{Direct SNN Training Approaches for Temporal Coding:} 
In the domain of TTFS coding, a convolutional-like coding method \cite{liu2018mtspike, 8297383} was proposed to directly train an SNN, which uses a temporal kernel to integrate temporal and spatial information. It can significantly reduce the model size and transform the spatial localities into temporal localities which can improve efficiency and accuracy. Some recent works use the surrogate gradient \cite{Liu2023-hh,zhang2020rectified} or surrogate model \cite{park2021training} to solve the non-differentiable backpropagation issue in deterministic neurons with temporal coding. 
Another technique is to directly train SNNs with stochastic neurons. The smoothing effect of stochastic neurons is crucial for enabling gradient-based optimization methods in SNNs \cite{neftci2019surrogate} by solving the non-linear, non-differentiable aspects of the spiking mechanism. Research by \cite{Training,Learning} introduced a stochastic neuron model for directly training SNNs. This model uses the generalized linear model (GLM) \cite{Pillow2005-gn} and first-to-spike coding. The GLM consists of a set of linear filters to process the incoming spikes, followed by a nonlinear function that computes the neuron's firing probability based on the filtered inputs. Subsequently, this model employs a stochastic process such as the Poisson process to generate spike trains.  However, the computational complexity of adapting a GLM for large-scale SNN training can be quite high. Other recent works have also explored stochastic SNNs with TTFS coding where the stochastic neuron is implemented by the intrinsic physics of spin devices \cite{yang2023leveraging}. However, existing research primarily focuses on shallow networks and MNIST-level datasets and lacks quantification of benefits offered by stochastic computing and temporal encoding in SNNs at scale. 

\section{Methods}
In this section, we introduce the methodology to train deterministic and stochastic SNNs with First-to-Spike coding where the key idea is to find the neuron that generates the first spike signal, thereby terminating the inference process. Associated loss function design and weight gradient calculations are also elaborated considering discontinuity issues observed in spiking neurons. 
\subsection{Deterministic SNN}\label{AA}
The Leaky Integrate-and-Fire (LIF) neuron model is one of the most recognized spiking neuron models in SNNs, primarily chosen for its balance between simplicity and biological plausibility \cite{1333071}. The LIF model simulates the behavior of neurons by accumulating input signals (voltage) until they reach the threshold. During this period, the accumulated voltage decays over time, which simulates the electrical resistance seen in real neuronal membranes. However, the process to generate a spike in LIF models is non-differentiable which makes it challenging for traditional gradient-based methods.
Defining a surrogate gradient (SG) as a continuous relaxation of the real gradients is one of the common ways to tackle the discontinuous spiking nonlinearity \cite{neftci2019surrogate}. 
The deterministic LIF neuron model used in our network can be summarized as:
\begin{equation}
V_i^t = \lambda V_i^{t-1} + \sum_{j}{w_{i,j}X_j^t} - (V_i^{t-1} \geq V_{th}) * V_{th}
\end{equation}
\noindent where, $V_i^t$ is the membrane potential of neuron $i$ at time $t$, $w_{i,j}$ is the weight connecting the pre-synapse neuron $j$ and post-synapse neuron $i$, $X_j^t$ is the input signal of pre-synapse neuron $j$ at time $t$, $V_{th}$ is the threshold, and $\lambda$ is the leak scaling factor. When neuron $i$'s membrane potential at time $t-1$ is larger than the threshold, it generates a spike and resets its membrane potential. Soft reset is used to reduce the membrane potential by the threshold instead of hard reset which resets the membrane potential to a certain value. This reset method ensures that the residual potential that exceeds the threshold is carried over to subsequent timesteps, thereby minimizing potential information loss \cite{Rathi2020Enabling}. 

The output spike train $o_i^t$ is generated by following this equation:
\begin{equation}
o_i^t = 
\begin{cases} 
1 & \text{if } V_i^t \geq V_{th} \\
0 & \text{if } V_i^t < V_{th}
\end{cases}
\label{eq:oit}
\end{equation}

The temporal cross-entropy loss function\cite{eshraghian2021training}, which integrates the principles of First-to-Spike coding, is formalized as follows:
For each neuron $i$, the estimated activation probability is computed using the equation:
\begin{equation}
p_i = \frac{e^{-t_i}}{\sum_{i=1}^n e^{-t_i}}
\label{eq:probability}
\end{equation}
where, $t_i$ is the time of the first spike of neuron $i$ and $n$ is the number of output neurons. The loss function is given by the following equation: 
\begin{equation}
L(\theta) = \sum_{i=1}^{n} y_i log(p_i)
\label{eq:loss}
\end{equation}
where, $y_i\in \{0,1\}$ is a one-hot target vector and $n$ is the number of output neurons. 
In the context of First-to-Spike coding, the goal is to minimize the time of the first spike of the correct neuron, which leads to maximizing its corresponding probability, as indicated by the cross-entropy loss function. 

The gradient of the weights corresponding to the deterministic LIF neuron model is given by the following equation:
\begin{equation}
\Delta w_{i,j} =\sum_t \frac{\partial L}{\partial t_i} \frac{\partial t_i}{\partial o_i^t} \frac{\partial o_i^t}{\partial V_i^t} \frac{\partial V_i^t}{\partial w_{i,j}} 
\end{equation}

Surrogate gradients provide a solution by approximating the gradient of the spike generation process, enabling gradient descent through these non-differentiable neurons. In the deterministic LIF neuron model, the process of extracting the first spike time from the output spike train is non-differentiable. Therefore, we use the sign estimator by replacing the gradient ${\partial t_i}/{\partial o_i^t}$  with $-1$ only at the time of the first spike \cite{eshraghian2021training}. Since Eqn. \ref{eq:oit} is also non-differentiable, in order to compute ${\partial o_i^t}/{\partial V_i^t}$, we need a surrogate gradient to solve the discontinuous spiking nonlinearity. In this paper, the $Arctan$ surrogate \cite{fang2021incorporating} is used. After employing the $Arctan$ surrogate, Eqn. \ref{eq:oit} can be written as:
\begin{equation}
o_i^t \approx \frac{1}{\pi} \arctan\left( \pi V_i^t \frac{\alpha}{2} \right)
\end{equation}

By using the surrogate gradient function, the discrete event is approximated as a differentiable function. The gradient ${\partial t_i}/{\partial o_i^t}$ can be expressed as:
\begin{equation}
\frac{\partial o_i^t}{\partial V_i^t} = \frac{1}{\pi} \frac{1}{1 + \left( \pi V_i^t \frac{\alpha}{2} \right)^2}
\label{eq:o/v}
\end{equation}

This allows the network to be trained using variants of backpropagation. In this paper, Backpropagation through time (BPTT) \cite{Wu_2018} is used where the network is unrolled across timesteps for backpropagation.

\subsection{Stochastic SNN}
On the other hand, the introduction of stochasticity can efficiently smoothen out discontinuous spiking nonlinearities. Inspired by \cite{Training}, we integrate stochastic LIF neurons with First-to-Spike coding. The membrane potential is computed by using the following equation: 
\begin{equation}
V_i^t = \frac{\lambda V_i^{t-1} + \sum_{j}{w_{i,j}X_j^t}}{k_i}
\end{equation}
where, $k_i$ is a scaling factor of the membrane potential of the neuron. Subsequently, the sigmoid activation function is used to calculate $p_i^t$, which is the probability of neuron $i$ generating a spike at time $t$. The probability $p_i^t$ is used to generate an independent and identically distributed (i.i.d.) Bernoulli value, which represents the discrete spike train generated by the neuron. Due to the non-differentiable nature of the Bernoulli function, it poses a problem for backpropagation techniques which rely on gradient-based optimization. To address this issue, we use the Straight-Through (ST) estimator \cite{DBLP}, which passes the gradient received from the deeper layer directly to the preceding layer without any modification in the backward phase. In the output layer, we compute the probability $P_t$ of the correct neuron to generate the earliest spike at time $t$ by the following equation:
\begin{equation}
P_t = p_c^t\prod_{i=1,i\neq c}^{n} \prod_{t'=1}^{t} (1-p_i^{t'})  \prod_{t'=1}^{t-1} (1-p_c^{t'})
\end{equation}
where, $p_c^t$ is the probability of correct neuron \textit{c} generating a spike at time $t$. This equation represents the probability that no wrong neurons generate a spike before the correct neuron produces a spike at time $t$.
We use the same ML (Maximum Likelihood) criterion used in \cite{Training} by maximizing the sum of all $P_t$s through the following equation:
\begin{equation}
L(\theta) = \log \left( \sum_{t=1}^{T} P_t \right)
\label{eq:loss2}
\end{equation}

 % However, the latency remains considerably high. A novel hybrid conversion was proposed by \cite{Rathi2020Enabling}, which uses a pre-trained ANN to initialize the network and employs BPTT with a surrogate gradient for further training. It has been proven to reduce timesteps while maintaining accuracy \cite{Rathi2020DIETSNNDI}. 
As the timestep increases, $P_t$ reduces, and the contributions to the overall losses diminish progressively which encourages neurons to fire earlier but not before the correct neuron, resulting in reduced latency. Furthermore, the BPTT algorithm is employed in a similar fashion as the deterministic SNN model, unfolding the network across timesteps and calculating the gradients of Eqn. \ref{eq:loss2} with respect to the weights at each timestep.

\section{Results}

In this section, we evaluate the accuracy, latency, sparsity, energy cost, and noise sensitivity of different types of models to evaluate the influence of information encoding, computing scheme, and training methods independently: ANN-SNN conversion utilizing deterministic neurons and rate coding (D-R-CONV) \cite{Exploring}, BPTT trained models with deterministic neurons utilizing rate coding (D-R-BPTT) \cite{Rathi2020Enabling}, deterministic neural networks trained by BPTT utilizing First-to-Spike coding (D-F-BPTT) and the stochastic neural networks trained by BPTT utilizing First-to-Spike coding (S-F-BPTT). It is worth noting that First-to-Spike coding is only applied in the final output layer to leverage the spike timing at which a neuron first spikes to encode information. Acronyms are used to simplify the naming that reflects its key features: The first part indicates the type of neurons: deterministic (D) and stochastic (S), the second part denotes the coding method: rate coding (R) and First-to-Spike coding (F), and the third part represents the training method: ANN-SNN conversion (CONV) and Backpropagation Through Time (BPTT). We will use these acronyms throughout the remainder of the paper for brevity. We conduct experiments for three neural network architectures, ranging from shallow to deep: 2-layer MLP, LeNet5, and VGG15.

\subsection{Datasets}

In this paper, we use the MNIST \cite{lecun-mnisthandwrittendigit-2010} and CIFAR-10 \cite{cifar} datasets for our experiments. In the preprocessing stage for the MNIST dataset, we adjust the pixel intensities from their original range of 0-255 to a normalized range of 0-1.  For the CIFAR-10 dataset, we use data augmentation to effectively increase the diversity of the training data and reduce overfitting. In our case, the random horizontal flipping is applied with a probability of 0.5, and the image is rotated at an angle randomly selected from a range of -15 to 15 degrees \cite{Shorten2019-go}. Our preprocessing also includes random cropping of images, with a padding of 4 pixels \cite{lee2014deeplysupervised}. To further augment the dataset, a random affine transformation is applied to the image. This includes shear-based transformations, where the degree of shear is precisely set to 10, effectively introducing a specific level of distortion to the images. Additionally, scaling adjustments are applied, altering the image size to fluctuate between 80\% and 120\% of the original size. The image attributes such as brightness, contrast, and saturation are adjusted \cite{cubuk2019autoaugment}, each by a factor of 0.2, to enhance model robustness against varying lighting and color conditions. Furthermore,  normalization of the input image data is employed based on the mean and standard deviation for each color channel in the CIFAR-10 dataset.
\subsection{Model Training}
\begin{table}[htbp]
  \centering
  \caption{Training Hyperparameters}
  \begin{tabular}{|l|c|c|c|c|}
    \hline
    \textbf{Dataset} & \multicolumn{2}{c|}{\textbf{MNIST}} & \multicolumn{2}{c|}{\textbf{CIFAR-10}} \\
    \hline
    Model & S-F-BPTT & D-F-BPTT & S-F-BPTT & D-F-BPTT \\
    \hline
    Leakage Factors & 0.7 & 0.9 & 0.7 & 0.9\\
    \hline
    Epoch& 150 & 150 & 1000 & 1000 \\
    \hline
    Batch Size & 512 & 512 & 64 & 64 \\
    \hline
    Learning Rate & 5e-2 & 1e-3 & 1e-2 & 5e-5 \\
    \hline
    Weight Decay & 1e-6 & 1e-4 & 1e-6 & 1e-2 \\
    \hline
    \makecell[l]{Scheduler \\ Step-Size} & 50 & 50 & 200 & 120 \\

    \hline
    \makecell[l]{Scheduler \\ Gamma}& 0.8 & 0.5 & 0.5 & 0.5 \\
    \hline
  \end{tabular}
  \label{hyperparameters}
\end{table}

\begin{table*}
  \centering
  \caption{Performance Comparison of Different Types of SNNs}
  \label{accuracy}
    \begin{tabular}{|c|c|c|c|c|c|}
    \hline
    \textbf{Dataset} & \textbf{Architecture} & \textbf{Model} & \textbf{Accuracy} & \textbf{Timesteps} & \textbf{Energy Cost}\\
    \hline
    \multirow{4}{*}{MNIST} & \multirow{4}{*}{2-Layer MLP} & D-R-CONV \cite{Exploring} & 97.74\% & 15 & 1.38\\
    \cline{3-6} 
    & &  D-R-BPTT \cite{Rathi2020Enabling}  & 98.57\% & 15 & 0.87\\
    \cline{3-6} 
    & &  S-F-BPTT & 98.62\% & \textbf{2.03} & 0.11\\
    \cline{3-6} 
    & &  D-F-BPTT & \textbf{98.76}\% & 3.10 & \textbf{0.07}\\
    \hline
   \multirow{4}{*}{MNIST} & \multirow{4}{*}{LeNet5} & D-R-CONV \cite{Exploring} & 97.74\% & 25 &1.44\\
    \cline{3-6}
    & &  D-R-BPTT \cite{Rathi2020Enabling} & 98.59\% & 35 & 4.95\\
    \cline{3-6}
    & &  S-F-BPTT & 98.69\% & \textbf{2.07} & 0.76\\
    \cline{3-6}
    & &  D-F-BPTT & \textbf{99.12}\% & 3.23 & \textbf{0.28}\\
    \hline
    \multirow{4}{*}{CIFAR-10} & \multirow{4}{*}{VGG15} & D-R-CONV \cite{Exploring} & \textbf{90.12}\% & 110 & 3.77\\
    \cline{3-6}
    & & D-R-BPTT \cite{Rathi2020Enabling} &89.61\% & 100 & 12.26\\
    \cline{3-6}
    & &  S-F-BPTT & 90.03\% & \textbf{5.43} & 0.81\\
    \cline{3-6}  
    & &  D-F-BPTT & 89.55\% &  10.95& \textbf{0.75}\\
    \hline
    \multicolumn{6}{|c|}{Other Temporal Coding Models} \\
    \hline
     \multirow{3}{*}{MNIST}
     & 2-Layer MLP& Sakemi et al. 2023 \cite{9490311} & 98.34\% & - & -\\
     \cline{2-6} 
     & 5Conv-1FC &  Kim et al. 2024 \cite{Rethinking} & 98.50\% & - & -\\   
    
    \cline{2-6}  
    & 2Conv-1FC & T2FSNN \cite{park2020t2fsnn} & 99.33\%& 40& -\\
    \hline
    \multirow{3}{*}{CIFAR-10}& \multirow{3}{*}{VGG16} & T2FSNN \cite{park2020t2fsnn} & 91.43\%& 680& -\\
    \cline{3-6}      
    & & Park et al. 2021 \cite{park2021training} & 91.90\% & 544 & -\\
    \cline{3-6}  
    & & DTA-TTFS \cite{Wei_2023_ICCV} & 93.05\% & 160 & - \\
    \hline
  \end{tabular}
\end{table*}

In the training of all architectures, the Adam optimizer \cite{kingma2017adam} is utilized, accompanied by a learning rate scheduler. For deterministic LIF neurons, the parameter $\alpha$ is set to 2 in Eqn \ref{eq:o/v}. The detailed hyperparameter settings are listed in Table \ref{hyperparameters}.
% \textcolor{blue}{The leakage factors for S-F-BPTT and D-F-BPTT are 0.7 and 0.9, respectively. The 2 Layer MLP and LeNet5 were trained on the MNIST dataset for 150 epochs with a batch size of 512. For the S-F-BPTT, the learning rate is 5e-2, with a weight decay of 1e-6, a scheduler step size of 50, and a gamma of 0.8. For D-F-BPTT, the learning rate is 1e-3 with a weight decay of 1e-4, and the scheduler has a step size of 50 and a gamma of 0.5. The VGG-15 was trained on the Cifar-10 dataset for 1000 epochs with a batch size of 64. In the case of the S-F-BPTT model, hyperparameters were set as follows: a learning rate of 0.01, a weight decay of 1e-6, and the scheduler with a step size of 200 and a gamma value of 0.5. For the D-F-BPTT model, the learning rate is 5e-5, the weight decay is 1e-2, and the step size and gamma of the scheduler are set to 120 and 0.5, respectively.}
Additionally, a critical aspect of SNN-specific optimization involves layerwise tuning of the neuron's firing threshold $V_{th}$ in the D-F-BPTT model and the scaling factor $k_i$ in the S-F-BPTT model (see Section III). For this purpose, we used a Neuroevolutionary optimized hybrid SNN training approach \cite{Lu2022-ki} where the trained model was subsequently optimized using the gradient-free differential evolution (DE) algorithm \cite{de, virtanen2020scipy} to achieve the best accuracy-latency tradeoff. Prior work has demonstrated that such a hybrid framework significantly outperforms approaches that combine such hyperparameter tuning during the BPTT training process itself \cite{Lu2022-ki}.

% D-R-BPTT, S-T-Conv, describe in text
\subsection{Quantitative Analysis}

\noindent \textbf{Accuracy:} The performance of each network is summarized in Table \ref{accuracy}. The accuracy is determined by calculating the mean value across ten independent runs. Transitioning from rate coding to temporal coding actually does not reduce the accuracy and even increases the accuracy in some cases. \textit{However, the introduction of stochasticity to the model causes a consistent increase in the network accuracy on the more complex CIFAR-10 dataset}. For complex datasets, the variability introduced by stochasticity could act as a form of data augmentation, presenting the network with a wider range of data during the training. This can prevent the model from overfitting, leading to better generalization. 

\vspace{2mm}

% Upon concluding the quantitative assessment of the model's accuracy, it is imperative to advance our analysis by incorporating the Fisher Information Matrix (FIM)\cite{Fisher1925TheoryOS}  which is used to quantify the amount of information about the training data contained in the model \cite{achille2019critical}, \cite{kim2022exploring},  \cite{lu2023deep}. The higher FIM indicates that the model can learn more information from training data and vice versa. The total FIM value (sum of the FIM of all learnable parameters on a logarithmic scale) of different neuron models in LeNet5 on the MNIST dataset is shown in Fig.~\ref{fim}. We note that the total FIM of S-F-BPTT is the highest, followed by the D-F-BPTT model, and the D-R-BPTT model is the lowest. It shows that the stochastic neuron with temporal coding can learn more information during the training process. This observation also explains why stochastic neurons utilizing temporal coding can achieve the highest levels of accuracy.
%Latency is the time that a network takes for an input to produce an output. In the SNNs, 
%latency is measured by the number of timesteps required for the network to respond to an input stimulus with the output spikes. 

\noindent \textbf{Inference Latency:} In SNNs, reducing latency without sacrificing accuracy can be a critical goal, allowing for faster and more energy-efficient computation. In particular, for the D-R-CONV and D-R-BPTT models, the optimal number of timesteps is determined by identifying the saturation point on a plot of timesteps versus accuracy, where further increase in timesteps no longer significantly improves model accuracy. For the S-F-BPTT and D-F-BPTT models, the inference latency is determined by averaging the number of timesteps at which the first spike is detected in the output layer across all input data. The differences in SNN inference latency in terms of timesteps are noted in Table \ref{accuracy}. \textit{Compared to other temporal coding models, First-to-Spike coding models show a significantly lower latency.} The First-to-Spike coding scheme requires only a single spike in the output layer to ascertain the result, reducing the number of timesteps significantly. On the other hand, rate coding relies on the frequency of spikes over time, and therefore the network needs a longer observation window to establish an accurate spike rate. This phenomenon is magnified when the dataset becomes complex. On the CIFAR-10 dataset, the rate coding approaches require a substantially higher number of timesteps to achieve the same level of accuracy compared to the models employing First-to-Spike coding.
%The results indicate that our models, namely the S-F-BPTT and D-F-BPTT models, use the First-to-Spike coding to reach the same level of accuracy with a smaller number of timesteps compared to the D-R-CONV and D-R-BPTT models which use rate coding. For the LeNet5 architecture on the MNIST dataset, the D-R-CONV and D-R-BPTT models require $7.7\times$ and $10.8\times$  timesteps compared to the D-F-BPTT. 
\textit{Interestingly, we find that the S-F-BPTT model reduces the latency even further in comparison to the D-F-BPTT model.} The stochastic nature of spike generation in S-F-BPTT models may cause an output spike generation even when the input stimulus is not too strong or the membrane potential is low, allowing for a faster response to the input. %However, this advantage is offset by a slight reduction in accuracy, as previously mentioned. 
\begin{figure*}[h]
    \subfigure[LeNet5 on MNIST Dataset]{
     \includegraphics[scale=0.22]
    {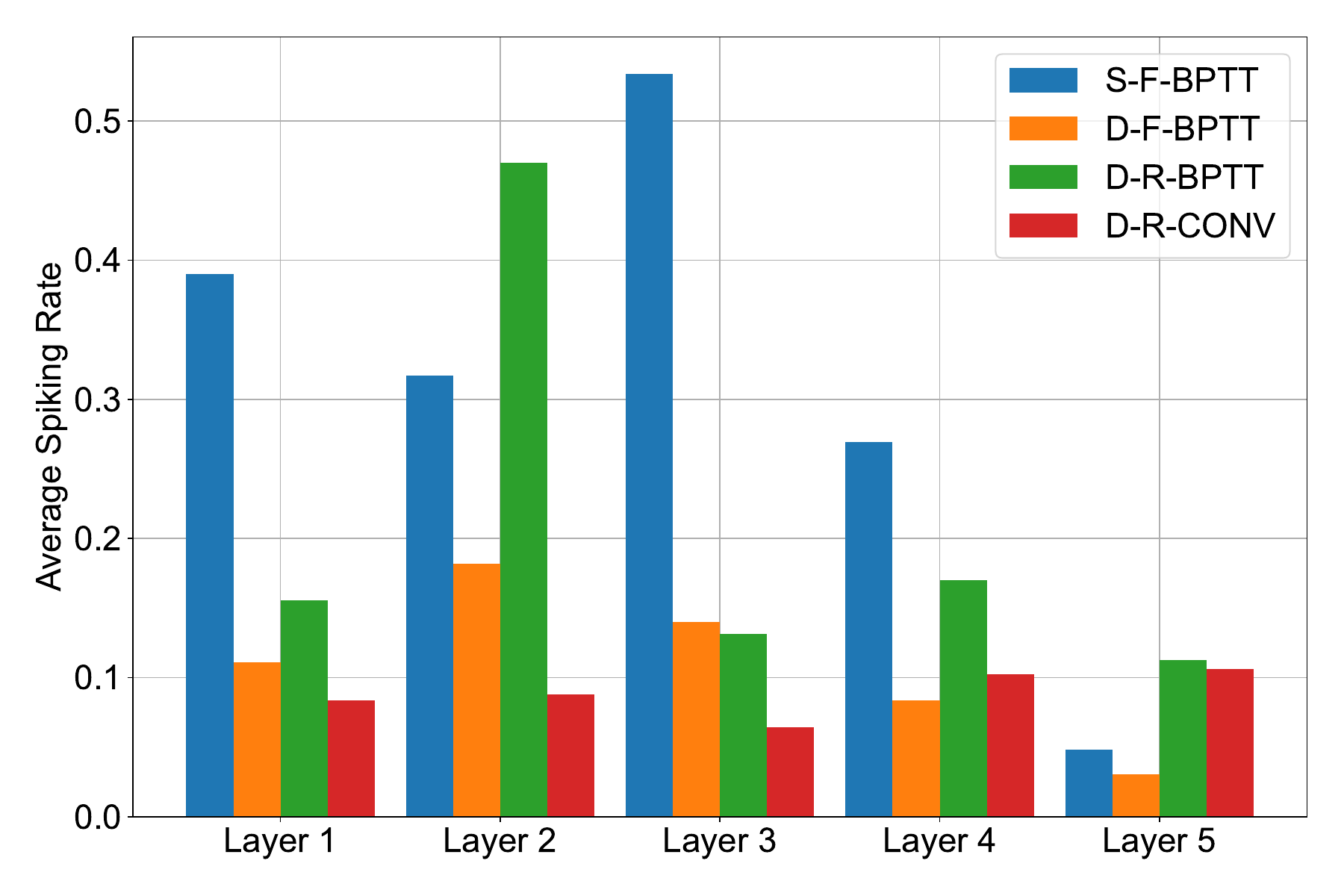}
    \label{sparsity_lenet5}}
    \subfigure[VGG15 on CIFAR-10 Dataset]{
    \includegraphics[scale=0.22]{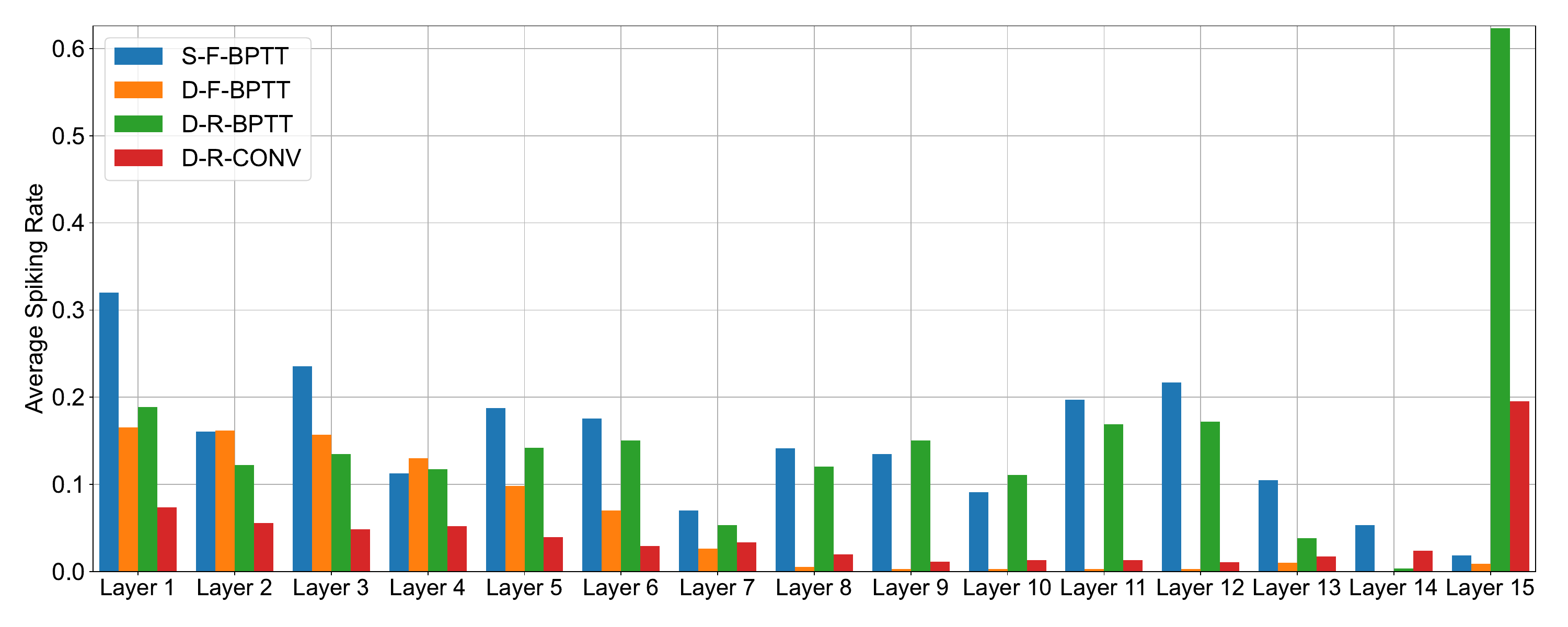}
    \label{sparsity_vgg15}}

    \caption{Average spiking rate of different neuronal layers for different SNN models (S-F-BPTT, D-F-BPTT, D-R-BPTT, D-R-CONV) corresponding to (a) LeNet5 on the MNIST dataset, (b) VGG15 on the CIFAR-10 dataset. 
    }
    \label{sparsity}
\end{figure*}
\vspace{2mm}

\noindent \textbf{Sparsity:} Spiking sparsity in SNNs is an important metric for evaluating the efficiency and functionality of models. The average spiking rate of a particular layer, defined as the average number of spikes that a neuron generates over a fixed time interval, is utilized to quantitatively represent sparsity. In this context, a higher average spike rate indicates lower sparsity,  and vice versa. The average spiking rate of each model across various layers for the LeNet5 architecture trained on the MNIST dataset and VGG15 architecture trained on the CIFAR-10 dataset is shown in Fig. \ref{sparsity}. \textit{Contrary to the common assumption of higher sparsity in temporal coding models than in rate coding models, the figure shows that first-to-spike models do show higher sparsity in the final layer, with the hidden layers presenting a contrary trend, especially for the stochastic model, as shown in the results}. The main reason is the necessity to encode the same information with reduced latency for temporally encoded models, which demands an increased spike count. It also explains why the S-F-BPTT model has the highest spiking rate with the lowest latency. Moreover, to achieve a reliable and consistent output in the presence of stochasticity, the stochastic SNN model needs to increase its spiking rate. This compensates for the unpredictability of individual spikes, ensuring that the overall signal transmission between neurons remains stable and correct.

% \subsection{Energy Cost}
\vspace{2mm}

\noindent \textbf{Energy Cost:} The total number of SNN computations, that serves as a proxy metric for the resultant energy consumption of the model when deployed on neuromorphic hardware \cite{Exploring}, is also a crucial factor in designing SNN models. The total ``energy cost" $E$ of each model, defined as the ratio of the number of computations performed in the SNN relative to that of an iso-architecture ANN, can be estimated as:
\begin{equation}
E= \sum_{i=2}^L{S_{i-1} \times T \times  \frac{{OP}_i}{\sum_{j=2}^L{{OP}_j}}}
\end{equation}
where, $L$ is the total number of layers, $S_{i-1}$ is the average spiking rate of the $(i-1)^{th}$ layer, $T$ is the number of timesteps used for inference, and ${OP}_i$ is the number of operations in the $i^{th}$ layer. Following \cite{Exploring}, ${OP}_i$ of convolutional layers and linear layers can be summarized as:
\begin{equation}
\begin{aligned}
{OP}_i = \begin{cases}
    C_I \cdot K_H \cdot K_W \cdot C_O \cdot O_H \cdot O_W& l_i \in \text{Conv} \\
    I_F \cdot O_F&l_i \in \text{Linear}
    
\end{cases}
\end{aligned}
\end{equation}
where, $l_i$ is the $i^{th}$ layer, $C_I$ and $C_O$ are the number of input and output channels, $K_H$ and $K_W$ are the height and width of the kernel, $O_H$ and $O_W$ are the height and width of the output, and $I_F$ and $O_F$ are the number of input and output features.
The results depicted in Table \ref{accuracy} demonstrate the energy cost of the four models: D-R-BPTT, D-R-CONV, S-F-BPTT, and D-F-BPTT. The D-R-BPTT model has the highest energy requirement, followed by the D-R-CONV, S-F-BPTT, and D-F-BPTT models, in descending order. %Specifically, the average energy consumption of the D-R-BPTT, D-R-CONV, and S-F-BPTT models is 17.5x, 5.09x, and 2.96x of the D-F-BPTT model's energy consumption, respectively.
\textit{The results demonstrate that the models using First-to-Spike coding are more energy efficient than rate-coded models. Furthermore, it is observed that the benefit of lower latency for the S-F-BPTT model is outweighed by its significantly higher spiking rate in contrast to the D-F-BPTT model, ultimately resulting in comparable or increased energy expenditure.} However, on the CIFAR-10 dataset, this difference is less pronounced, as the D-F-BPTT model requires almost twice the number of timesteps compared to the S-F-BPTT model, resulting in only a slight difference in the total energy cost.

\vspace{2mm}
\begin{figure}[ht]
    \centering
    
    \includegraphics[scale=0.3]{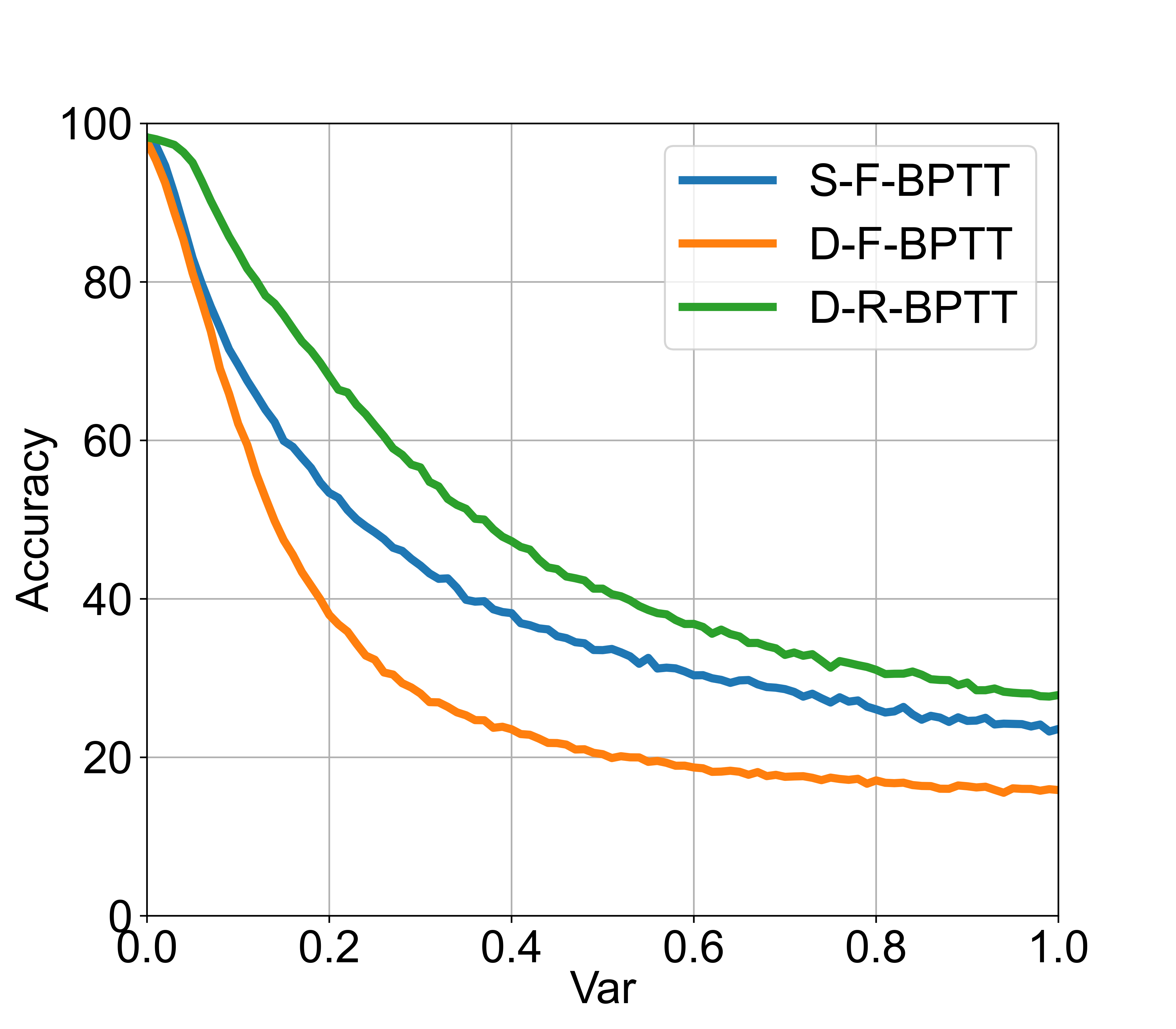}
    \label{fig:Gaussian}
    % \subfigure[Salt Noise]{
    % \includegraphics[scale=0.3]{figures/noise/salt_noise.png}
    % \label{fig:salt}}
    % \subfigure[Salt-and-Pepper Noise]{
    % \includegraphics[scale=0.3]{figures/noise/salt&pepper_noise.png}
    % \label{fig:salt_pepper}}
\caption{Comparative analysis of the performance of different SNN models (S-F-BPTT, D-F-BPTT, D-R-BPTT) under Gaussian noise.}
\label{noise}
\end{figure}
\noindent \textbf{Noise Sensitivity:}
A key aspect of ML model design is to ensure robustness to noise. %A common strategy for evaluating the noise robustness of SNNs is to introduce noise in the input data. This method provides a controlled way to simulate the impact of real-world noise conditions on the network. 
A model's noise sensitivity can be measured by adding different levels of noise to the input and observing the impact on the network's accuracy. 
%We introduce noise to the data and assess the impact on model accuracy to measure sensitivity. 
%A model that shows substantial performance decline with slight noise is considered overly sensitive, while one that sustains performance despite significant noise demonstrates robustness. 
In this paper, Gaussian noise is used to assess how well the model can maintain its performance under noisy conditions. The variance of Gaussian noise is adjusted from 0 to 1.
%These three types of noise are added to the test data respectively and the changes in accuracy are measured to evaluate the sensitivity of the D-F-BPTT, S-F-BPTT, and D-R-BPTT models. 
Fig.~\ref{noise} shows the relationship between accuracy degradation and the magnitude of applied noise. It can be observed that the D-R-BPTT model demonstrates a higher tolerance to noise, maintaining higher accuracy as the noise intensity increases. Since it uses rate coding, which encodes information by the frequency of multiple spikes over time, individual perturbations caused by noise have less impact on the overall information conveyed. \textit{Temporal coding models (D-F-BPTT and S-F-BPTT) are more sensitive to noise because they rely on the precise timing of spikes to encode information. However, the stochastic model has better performance than the deterministic model at high noise levels. This can be attributed to the stochasticity which is incorporated during the training process itself and therefore can provide more resilience to noise (through more tolerance towards the precision of individual spikes).}

% \subsection{CIFAR-10}
%  The average top-10 accuracy for the S-F-BPTT model is 90.03, while for the D-F-BPTT model, it is ***. Specifically, the D-R-CONV model necessitates the most timesteps, totaling 110, with the D-R-BPTT model close behind at 100 timesteps. Conversely, the S-F-BPTT and D-F-BPTT models demonstrate significantly lower latency, recorded at 5.94 and **, respectively.

\section{Conclusions}
In summary, our research explores the interplay of deterministic/stochastic computing with First-to-Spike information coding in SNNs. This integration bridges a gap in current research, demonstrating scalable direct training of SNNs with temporal encoding on large-scale datasets and deep architectures. We showcase that First-to-Spike coding has significant performance benefits for SNN architectures in contrast to traditional rate-based models with regard to various metrics, including latency, sparsity, and energy efficiency.
%One of the key challenges in scaling up stochastic models with temporal coding schemes is the precision and synchronization needed for effective temporal coding increase. Moving from simpler tasks, like those using the MNIST dataset, to more complex ones, such as the CIFAR-10 dataset, introduces greater complexity in maintaining effective temporal coding. 
% This is due to the necessity for more precise timing control over spike generation, a requirement that grows with the network's complexity and the sophistication of the tasks it undertakes. 
We also underscore notable trade-offs between the stochastic and deterministic SNN models in temporal encoding scenarios. Stochastic models reduce latency and provide enhanced noise robustness, which is important for real-time confidence-critical applications.  However, this advantage is offered at the expense of a slight decrement in sparsity, which consequentially results in higher energy costs compared to deterministic SNNs employing First-to-Spike coding. In terms of accuracy, stochastic SNNs have the potential to aid in better generalization, especially for complex datasets.
%These trade-offs highlight the considerations necessary for optimizing neuromorphic systems, aiming to balance latency, accuracy, sparsity, energy efficiency, and robustness.  This research establishes a foundation for future progress in stochastic SNNs utilizing temporal coding. 
Although our results are promising, scaling this method to ImageNet level vision tasks as well as beyond vision applications could be a future research direction. Energy and sparsity aware training techniques can be also considered for stochastic SNN models with temporal encoding to further enhance its applicability for resource-constrained edge devices.

\section*{Acknowledgments}
This material is based upon work supported in part by the U.S. Department of Energy, Office of Science, Office of Advanced Scientific Computing Research, under Award Number \#DE-SC0021562, the U.S. National Science Foundation under award No. CCSS \#2333881, CCF \#1955815, CAREER \#2337646 and EFRI BRAID \#2318101 and by Oracle Cloud credits and related resources provided by the Oracle for Research program.
\vspace{12pt}

%\bibliography{cites}

\begin{thebibliography}{10}
\providecommand{\url}[1]{#1}
\csname url@samestyle\endcsname
\providecommand{\newblock}{\relax}
\providecommand{\bibinfo}[2]{#2}
\providecommand{\BIBentrySTDinterwordspacing}{\spaceskip=0pt\relax}
\providecommand{\BIBentryALTinterwordstretchfactor}{4}
\providecommand{\BIBentryALTinterwordspacing}{\spaceskip=\fontdimen2\font plus
\BIBentryALTinterwordstretchfactor\fontdimen3\font minus \fontdimen4\font\relax}
\providecommand{\BIBforeignlanguage}[2]{{%
\expandafter\ifx\csname l@#1\endcsname\relax
\typeout{** WARNING: IEEEtran.bst: No hyphenation pattern has been}%
\typeout{** loaded for the language `#1'. Using the pattern for}%
\typeout{** the default language instead.}%
\else
\language=\csname l@#1\endcsname
\fi
#2}}
\providecommand{\BIBdecl}{\relax}
\BIBdecl

\bibitem{10.1162/neco_a_01499}
A.~Javanshir, T.~T. Nguyen, M.~A.~P. Mahmud, and A.~Z. Kouzani, ``{Advancements in Algorithms and Neuromorphic Hardware for Spiking Neural Networks},'' \emph{Neural Computation}, vol.~34, no.~6, pp. 1289--1328, 05 2022.

\bibitem{10.5555/281543.281637}
W.~Maas, ``Networks of spiking neurons: The third generation of neural network models,'' \emph{Trans. Soc. Comput. Simul. Int.}, vol.~14, no.~4, p. 1659–1671, dec 1997.

\bibitem{Guo2021NeuralCI}
W.~Guo, M.~E. Fouda, A.~M. Eltawil, and K.~N. Salama, ``Neural coding in spiking neural networks: A comparative study for robust neuromorphic systems,'' \emph{Frontiers in Neuroscience}, vol.~15, 2021.

\bibitem{Retina}
T.~Gollisch and M.~Meister, ``Rapid neural coding in the retina with relative spike latencies,'' \emph{Science}, vol. 319, no. 5866, pp. 1108--1111, 2008.

\bibitem{HEIL2004461}
P.~Heil, ``First-spike latency of auditory neurons revisited,'' \emph{Current Opinion in Neurobiology}, vol.~14, no.~4, pp. 461--467, 2004.

\bibitem{stanojevic2022exact}
A.~Stanojevic, S.~Woźniak, G.~Bellec, G.~Cherubini, A.~Pantazi, and W.~Gerstner, ``An exact mapping from relu networks to spiking neural networks,'' 2022.

\bibitem{8351295}
B.~Rueckauer and S.-C. Liu, ``Conversion of analog to spiking neural networks using sparse temporal coding,'' in \emph{2018 IEEE International Symposium on Circuits and Systems (ISCAS)}, 2018, pp. 1--5.

\bibitem{lin2024benchmarking}
J.~Lin, S.~Lu, M.~Bal, and A.~Sengupta, ``Benchmarking spiking neural network learning methods with varying locality,'' \emph{arXiv preprint arXiv:2402.01782}, 2024.

\bibitem{bal2024spikingbert}
M.~Bal and A.~Sengupta, ``Spikingbert: Distilling bert to train spiking language models using implicit differentiation,'' in \emph{Proceedings of the AAAI Conference on Artificial Intelligence}, vol.~38, no.~10, 2024, pp. 10\,998--11\,006.

\bibitem{zhu2023spikegpt}
R.-J. Zhu, Q.~Zhao, G.~Li, and J.~K. Eshraghian, ``Spikegpt: Generative pre-trained language model with spiking neural networks,'' \emph{arXiv preprint arXiv:2302.13939}, 2023.

\bibitem{Rathi2020Enabling}
N.~Rathi, G.~Srinivasan, P.~Panda, and K.~Roy, ``Enabling deep spiking neural networks with hybrid conversion and spike timing dependent backpropagation,'' in \emph{International Conference on Learning Representations}, 2020.

\bibitem{Kaiser_2020}
\BIBentryALTinterwordspacing
J.~Kaiser, H.~Mostafa, and E.~Neftci, ``Synaptic plasticity dynamics for deep continuous local learning (decolle),'' \emph{Frontiers in Neuroscience}, vol.~14, May 2020. [Online]. Available: \url{http://dx.doi.org/10.3389/fnins.2020.00424}
\BIBentrySTDinterwordspacing

\bibitem{scellier2017equilibrium}
B.~Scellier and Y.~Bengio, ``Equilibrium propagation: Bridging the gap between energy-based models and backpropagation,'' 2017.

\bibitem{bal2022sequence}
M.~Bal and A.~Sengupta, ``Sequence learning using equilibrium propagation,'' \emph{arXiv preprint arXiv:2209.09626}, 2022.

\bibitem{lu2023deep}
S.~Lu and A.~Sengupta, ``{Deep Unsupervised Learning Using Spike-Timing-Dependent Plasticity},'' \emph{Neuromorphic Computing and Engineering}, 2023.

\bibitem{Gerstner1993-hc}
W.~Gerstner, R.~Ritz, and J.~L. van Hemmen, ``\BIBforeignlanguage{en}{Why spikes? hebbian learning and retrieval of time-resolved excitation patterns},'' \emph{\BIBforeignlanguage{en}{Biol. Cybern.}}, vol.~69, no. 5-6, pp. 503--515, Sep. 1993.

\bibitem{lapicque1907recherches}
L.~Lapicque, ``Recherches quantitatives sur l’excitation électrique des nerfs traitée comme une polarisation,'' \emph{J Physiol Paris}, vol.~9, pp. 620--635, 1907.

\bibitem{7329679}
W.~Maass, ``To spike or not to spike: That is the question,'' \emph{Proceedings of the IEEE}, vol. 103, no.~12, pp. 2219--2224, 2015.

\bibitem{neftci2019surrogate}
E.~O. Neftci, H.~Mostafa, and F.~Zenke, ``Surrogate gradient learning in spiking neural networks,'' 2019.

\bibitem{9605019}
A.~Ardakani, A.~Ardakani, and W.~J. Gross, ``Fault-tolerance of binarized and stochastic computing-based neural networks,'' in \emph{2021 IEEE Workshop on Signal Processing Systems (SiPS)}, 2021, pp. 52--57.

\bibitem{hinton2012improving}
G.~E. Hinton, N.~Srivastava, A.~Krizhevsky, I.~Sutskever, and R.~R. Salakhutdinov, ``Improving neural networks by preventing co-adaptation of feature detectors,'' 2012.

\bibitem{Sengupta_2016}
A.~Sengupta, M.~Parsa, B.~Han, and K.~Roy, ``Probabilistic deep spiking neural systems enabled by magnetic tunnel junction,'' \emph{IEEE Transactions on Electron Devices}, vol.~63, no.~7, p. 2963–2970, Jul. 2016.

\bibitem{Yang_2020}
K.~Yang and A.~Sengupta, ``Stochastic magnetoelectric neuron for temporal information encoding,'' \emph{Applied Physics Letters}, vol. 116, no.~4, Jan. 2020.

\bibitem{islam2023hardware}
A.~Islam, K.~Yang, A.~K. Shukla, P.~Khanal, B.~Zhou, W.-G. Wang, and A.~Sengupta, ``Hardware in loop learning with spin stochastic neurons,'' \emph{arXiv preprint arXiv:2305.03235}, 2023.

\bibitem{islam2023hybrid}
A.~Islam, A.~Saha, Z.~Jiang, K.~Ni, and A.~Sengupta, ``Hybrid stochastic synapses enabled by scaled ferroelectric field-effect transistors,'' \emph{Applied Physics Letters}, vol. 122, no.~12, 2023.

\bibitem{Ma2023}Ma, G., Yan, R. \& Tang, H. Exploiting noise as a resource for computation and learning in spiking neural networks. {\em Patterns}, vol. 4, no. 10, pp. 100831, 2023.

\bibitem{Training}
A.~Bagheri, O.~Simeone, and B.~Rajendran, ``Training probabilistic spiking neural networks with first-to-spike decoding,'' 2018.

\bibitem{Learning}
B.~Rosenfeld, O.~Simeone, and B.~Rajendran, ``Learning first-to-spike policies for neuromorphic control using policy gradients,'' in \emph{2019 IEEE 20th International Workshop on Signal Processing Advances in Wireless Communications (SPAWC)}, 2019, pp. 1--5.

\bibitem{yang2023leveraging}
K.~Yang, D.~P. Gm, and A.~Sengupta, ``Leveraging probabilistic switching in superparamagnets for temporal information encoding in neuromorphic systems,'' \emph{IEEE Transactions on Computer-Aided Design of Integrated Circuits and Systems}, 2023.

\bibitem{lecun-mnisthandwrittendigit-2010}
Y.~LeCun and C.~Cortes, ``{MNIST} handwritten digit database,'' 2010.

\bibitem{cifar}
A.~Krizhevsky, V.~Nair, and G.~Hinton, ``Cifar-10 (canadian institute for advanced research).''

\bibitem{gerstner_kistler_2002}
W.~Gerstner and W.~M. Kistler, \emph{Spiking Neuron Models: Single Neurons, Populations, Plasticity}.\hskip 1em plus 0.5em minus 0.4em\relax Cambridge University Press, 2002.

\bibitem{G_ltz_2021}
J.~Göltz, L.~Kriener, A.~Baumbach, S.~Billaudelle, O.~Breitwieser, B.~Cramer, D.~Dold, A.~F. Kungl, W.~Senn, J.~Schemmel, K.~Meier, and M.~A. Petrovici, ``Fast and energy-efficient neuromorphic deep learning with first-spike times,'' \emph{Nature Machine Intelligence}, vol.~3, no.~9, p. 823–835, Sep. 2021.

\bibitem{9706185}
S.~Oh, D.~Kwon, G.~Yeom, W.-M. Kang, S.~Lee, S.~Y. Woo, J.~Kim, and J.-H. Lee, ``Neuron circuits for low-power spiking neural networks using time-to-first-spike encoding,'' \emph{IEEE Access}, vol.~10, pp. 24\,444--24\,455, 2022.

\bibitem{park2020t2fsnn}
S.~Park, S.~Kim, B.~Na, and S.~Yoon, ``T2fsnn: Deep spiking neural networks with time-to-first-spike coding,'' 2020.

\bibitem{Rethinking}
\BIBentryALTinterwordspacing
Y.~Kim, A.~Kahana, R.~Yin, Y.~Li, P.~Stinis, G.~E. Karniadakis, and P.~Panda, ``Rethinking skip connections in spiking neural networks with time-to-first-spike coding,'' \emph{Frontiers in Neuroscience}, vol.~18, 2024. [Online]. Available: \url{https://www.frontiersin.org/journals/neuroscience/articles/10.3389/fnins.2024.1346805}
\BIBentrySTDinterwordspacing

\bibitem{Liu2023-hh}
S.~Liu, V.~C.~H. Leung, and P.~L. Dragotti, ``\BIBforeignlanguage{en}{First-spike coding promotes accurate and efficient spiking neural networks for discrete events with rich temporal structures},'' \emph{\BIBforeignlanguage{en}{Front. Neurosci.}}, vol.~17, p. 1266003, Oct. 2023.

\bibitem{Wei_2023_ICCV}
W.~Wei, M.~Zhang, H.~Qu, A.~Belatreche, J.~Zhang, and H.~Chen, ``Temporal-coded spiking neural networks with dynamic firing threshold: Learning with event-driven backpropagation,'' in \emph{Proceedings of the IEEE/CVF International Conference on Computer Vision (ICCV)}, October 2023, pp. 10\,552--10\,562.

\bibitem{park2021training}
S.~Park and S.~Yoon, ``Training energy-efficient deep spiking neural networks with time-to-first-spike coding,'' 2021.

\bibitem{Mozafari_2018}
M.~Mozafari, S.~R. Kheradpisheh, T.~Masquelier, A.~Nowzari-Dalini, and M.~Ganjtabesh, ``First-spike-based visual categorization using reward-modulated stdp,'' \emph{IEEE Transactions on Neural Networks and Learning Systems}, vol.~29, no.~12, p. 6178–6190, Dec. 2018.

\bibitem{Exploring}
S.~Lu and A.~Sengupta, ``Exploring the connection between binary and spiking neural networks,'' \emph{Frontiers in Neuroscience}, vol.~14, jun 2020.

\bibitem{sengupta2019going}
A.~Sengupta, Y.~Ye, R.~Wang, C.~Liu, and K.~Roy, ``Going deeper in spiking neural networks: Vgg and residual architectures,'' 2019.

\bibitem{li2021free}
Y.~Li, S.~Deng, X.~Dong, R.~Gong, and S.~Gu, ``A free lunch from ann: Towards efficient, accurate spiking neural networks calibration,'' 2021.

\bibitem{liu2018mtspike}
T.~Liu, Z.~Liu, F.~Lin, Y.~Jin, G.~Quan, and W.~Wen, ``Mt-spike: A multilayer time-based spiking neuromorphic architecture with temporal error backpropagation,'' 2018.

\bibitem{8297383}
T.~Liu, L.~Jiang, Y.~Jin, G.~Quan, and W.~Wen, ``Pt-spike: A precise-time-dependent single spike neuromorphic architecture with efficient supervised learning,'' in \emph{2018 23rd Asia and South Pacific Design Automation Conference (ASP-DAC)}, 2018, pp. 568--573.

\bibitem{zhang2020rectified}
M.~Zhang, J.~Wang, B.~Amornpaisannon, Z.~Zhang, V.~Miriyala, A.~Belatreche, H.~Qu, J.~Wu, Y.~Chua, T.~E. Carlson, and H.~Li, ``Rectified linear postsynaptic potential function for backpropagation in deep spiking neural networks,'' 2020.

\bibitem{Pillow2005-gn}
J.~W. Pillow, L.~Paninski, V.~J. Uzzell, E.~P. Simoncelli, and E.~J. Chichilnisky, ``\BIBforeignlanguage{en}{Prediction and decoding of retinal ganglion cell responses with a probabilistic spiking model},'' \emph{\BIBforeignlanguage{en}{J. Neurosci.}}, vol.~25, no.~47, pp. 11\,003--11\,013, Nov. 2005.

\bibitem{1333071}
E.~Izhikevich, ``Which model to use for cortical spiking neurons?'' \emph{IEEE Transactions on Neural Networks}, vol.~15, no.~5, pp. 1063--1070, 2004.

\bibitem{eshraghian2021training}
J.~K. Eshraghian, M.~Ward, E.~Neftci, X.~Wang, G.~Lenz, G.~Dwivedi, M.~Bennamoun, D.~S. Jeong, and W.~D. Lu, ``Training spiking neural networks using lessons from deep learning,'' \emph{Proceedings of the IEEE}, vol. 111, no.~9, pp. 1016--1054, 2023.

\bibitem{fang2021incorporating}
W.~Fang, Z.~Yu, Y.~Chen, T.~Masquelier, T.~Huang, and Y.~Tian, ``Incorporating learnable membrane time constant to enhance learning of spiking neural networks,'' 2021.

\bibitem{Wu_2018}
Y.~Wu, L.~Deng, G.~Li, J.~Zhu, and L.~Shi, ``Spatio-temporal backpropagation for training high-performance spiking neural networks,'' \emph{Frontiers in Neuroscience}, vol.~12, May 2018.

\bibitem{DBLP}
Y.~Bengio, N.~L{\'{e}}onard, and A.~C. Courville, ``Estimating or propagating gradients through stochastic neurons for conditional computation,'' \emph{CoRR}, vol. abs/1308.3432, 2013.

\bibitem{Shorten2019-go}
C.~Shorten and T.~M. Khoshgoftaar, ``\BIBforeignlanguage{en}{A survey on image data augmentation for deep learning},'' \emph{\BIBforeignlanguage{en}{J. Big Data}}, vol.~6, no.~1, Dec. 2019.

\bibitem{lee2014deeplysupervised}
C.-Y. Lee, S.~Xie, P.~Gallagher, Z.~Zhang, and Z.~Tu, ``Deeply-supervised nets,'' 2014.

\bibitem{cubuk2019autoaugment}
E.~D. Cubuk, B.~Zoph, D.~Mane, V.~Vasudevan, and Q.~V. Le, ``Autoaugment: Learning augmentation policies from data,'' 2019.

\bibitem{kingma2017adam}
D.~P. Kingma and J.~Ba, ``Adam: A method for stochastic optimization,'' 2017.

\bibitem{Lu2022-ki}
S.~Lu and A.~Sengupta, ``\BIBforeignlanguage{en}{Neuroevolution guided hybrid spiking neural network training},'' \emph{\BIBforeignlanguage{en}{Front. Neurosci.}}, vol.~16, p. 838523, Apr. 2022.

\bibitem{de}
R.~Storn and K.~Price, ``Differential evolution - a simple and efficient heuristic for global optimization over continuous spaces,'' \emph{Journal of Global Optimization}, vol.~11, pp. 341--359, 01 1997.

\bibitem{virtanen2020scipy}
P.~Virtanen, R.~Gommers, T.~E. Oliphant, M.~Haberland, T.~Reddy, D.~Cournapeau, E.~Burovski, P.~Peterson, W.~Weckesser, J.~Bright \emph{et~al.}, ``{SciPy} 1.0: fundamental algorithms for scientific computing in python,'' \emph{Nature methods}, vol.~17, no.~3, pp. 261--272, 2020.

\bibitem{9490311}
Y.~Sakemi, K.~Morino, T.~Morie, and K.~Aihara, ``A supervised learning algorithm for multilayer spiking neural networks based on temporal coding toward energy-efficient vlsi processor design,'' \emph{IEEE Transactions on Neural Networks and Learning Systems}, vol.~34, no.~1, pp. 394--408, 2023.

\end{thebibliography}
%\bibliographystyle{IEEEtran}


\end{document}